\documentclass{article}
\usepackage[T1]{fontenc}

\usepackage{graphicx}
\usepackage{authblk}
\usepackage{color}
\usepackage{doi}
\usepackage{amsmath, amssymb} 
\usepackage{caption}   
\usepackage{hyperref}
\usepackage{tikz}
\usetikzlibrary{arrows.meta,backgrounds}

\pgfdeclarelayer{nodelayer}
\pgfdeclarelayer{edgelayer}
\pgfsetlayers{background,main,nodelayer,edgelayer}

\begin{document}
\title{A Spectral Interpretation of Redundancy in a Graph Reservoir}

\author{
  Anna Bison and Alessandro Sperduti
}
\date{}
\affil{Department of Mathematics ``Tullio Levi-Civita''\\ University of Padova, Padova, Italy \\ 

\texttt{anna.bison@studenti.unipd.it, alessandro.sperduti@unipd.it}}
\maketitle             
\begin{abstract}
Reservoir computing has been successfully applied to graphs as a preprocessing method to improve the training efficiency of Graph Neural Networks (GNNs). However, a common issue that arises when repeatedly applying layer operators on graphs is over-smoothing, which consists in the convergence of graph signals toward low-frequency components of the graph Laplacian.
This work revisits the definition of the reservoir in the Multiresolution Reservoir Graph Neural Network (MRGNN), a spectral reservoir model, and proposes a variant based on a Fairing algorithm originally introduced in the field of surface design in computer graphics. This algorithm provides a pass-band spectral filter that allows smoothing without shrinkage, and it can be adapted to the graph setting through the Laplacian operator.
Given its spectral formulation, this method naturally connects to GNN architectures for tasks where smoothing, when properly controlled, can be beneficial,such as graph classification. The core contribution of the paper lies in the theoretical analysis of the algorithm from a random walks perspective. In particular, it shows how tuning the spectral coefficients can be interpreted as modulating the contribution of redundant random walks. Exploratory experiments based on the MRGNN architecture illustrate the potential of this approach and suggest promising directions for future research.\\

\newcommand{\keywords}[1]{\noindent\textbf{Keywords:} #1}
\keywords{Reservoir Computing $\cdot$ Fairing algorithm $\cdot$ GNN}
\end{abstract}

\section{Introduction}

In recent years, Reservoir Computing (RC) has been successfully applied to learning in graph domains in order to reduce the training computational burden while maintaining excellent performances (see, for example, \cite{WangLWZCDWZLGXLCWSL23,BianchiGM22,pasa2021multiresolution}). In almost all the proposed RC models, the reservoir is based on convolution operators for graphs (e.g., the ones defined in~\cite{kipf2017gcn} for Graph Convolutional Neural Networks) that inevitably lead to a smoothing effect on the embeddings of the nodes of the processed graphs.  
The smoothing effect can be a desirable property in tasks such as graph classification, since it filters node signals that have high local variation, often encoding local information, and selects low-frequency components, which are more likely to encode global information that is more important for this kind of task. For this reason, it could make sense to implement an algorithm that optimizes smoothing in a model that performs graph classification. 
Indeed, the main risk when dealing with smoothing filters for GNNs is the over-smoothing problem, extensively discussed in~\cite{rusch2023survey}, which consists in the exponential convergence of the node embeddings of a graph to the same vector, or to embeddings whose differences only depend on the node-degree distribution.
In recent literature, some methods have been proposed to manage over-smoothing by acting on the input expression without altering the model definition, as in~\cite{zhao2020pairnorm}. These kinds of methods may not solve the problem in cases where GNN layers are defined as low-pass filters, since they will behave like that independently of the input shape.

In this work, we explore, mainly from a theoretical point of view, how the Fairing algorithm introduced in~\cite{taubin1995smoothing} relates to spectral-based GNNs, and in particular to the \textit{Multiresolution Reservoir Graph Neural Network} (MRGNN), introduced in~\cite{pasa2021multiresolution}. 
The main motivation for considering the Fairing algorithm hinges on the fact that it proposes a procedure to define a filter that selects not only the desired frequencies to dampen or amplify, but also the number of iterations needed to reach the desired result. Even though it was proposed almost twenty years before the definition of Graph Convolutional Networks (GCN)~\cite{kipf2017gcn}, it effectively explains how to perform the equivalent of a convolution of degree-$n$, obtained by iterating $n$ times a layer that is a degree-one polynomial of a spectral operator (which, in the case of graphs, can be chosen to be the symmetric Laplacian). The difference is that in the case of the Fairing algorithm, the parameters are tuned by the user, who, being familiar with spectral methods for signal processing, knows what is needed. In this work, these parameters will be presented from the perspective of random walks, as equivalent to controlling the weight with which specific walks contribute to the diffusion of signals through graph nodes. Based on this theoretical analysis, we propose a novel Fairing-based reservoir that we preliminary experiment within the MRGNN architecture, obtaining encouraging results.

\section{Background}
Here we concisely present the definitions of the two conceptual assets that form the basis of our analysis. The MRGNN model~\cite{pasa2021multiresolution}, focusing on reservoir computing, and the Fairing algorithm~\cite{taubin1995smoothing}. Despite their differences, they can be linked via a spectral perspective.

\subsection {\bf Notation}
Uppercase letters will refer to matrices, and the elements of a matrix $X$ are indicated as $[X]_{ij}=x_{ij}$.
Let $G(V, E, (X, A))$ be a graph, where $V$ denotes the set of nodes, $E$ denotes the set of edges. Nodes are referred to by their indices, i.e., $i \in V$ denotes the $i$-th node, and $(i, j) \in E$ indicates that there is an edge connecting node $i$ and node $j$. $X$ denotes the embedding matrix: its $i$-th row denotes features vector (embedding) of node $i$, and the $j$-th column denotes the graph scalar signal associated to the $j$-th feature. $A$ denotes the adjacency matrix, whose elements $a_{ij}= 1 \iff (i, j) \in E$. $D$ denotes the diagonal degree matrix. Let $\Delta_{\text{sym}}=I -D^{-1/2}AD^{-1/2}$ be the symmetric normalized Laplacian. The set of first neighbours of node $i$ is denoted with $\mathcal{N}_i$.

\subsection {\bf Multiresolution Reservoir Graph Neural Network}
The MRGNN is a supervised graph neural network model based on a variant of a reservoir computing model, a $k$-hop reservoir that is mathematically equivalent to evolve the input embedding matrix with a pass-band filter.
In a typical RC model, recursive neurons are run until a precise condition is satisfied (either convergence or maximum number of iterations). 
In MRGNN instead, the reservoir consists in a forward non-linear model that extracts graph features up to the $k$-hop neighbourhood, called \textit{multiresolution features}.
Then, intermediate features extracted for all the neighbourhoods of intermediate levels up to the $k$-th are considered simultaneously, and linearly combined with a method that presents some analogies with a spectral approach. Specifically, the formal definition of the transformation implemented by the reservoir consists in $k$ iterations of a convolution:
\[
H^{k, \mathcal{T}} =
\begin{bmatrix}
X ,\sigma(\tilde{A}X),  \sigma(\tilde{A} \sigma(\tilde{A}X) ), \dots
\end{bmatrix}=
\begin{bmatrix}
H^{\;k, \mathcal{T}}_{(0)}, H^{\;k, \mathcal{T}}_{(1)}, H^{\;k, \mathcal{T}}_{(2)}, \dots, H^{\;k, \mathcal{T}}_{(k)}
\end{bmatrix}
\]
where $\sigma$ is the tanh activation function, $\tilde{A}= \mathcal{T}(A)$ is a function of the adjacency matrix and  $H^{\;k, \mathcal{T}}_{(i)}$ represents the embedding matrix of node representations where the $j$-th one is obtained by convolving signals from nodes connected through a path of length $i$ from the $j$-th node.

Finally, a readout function is implemented taking in input these multiresolution features and performing the final classification or regression task. In the paper, different definitions of the readout function, spanning from linear model to neural network are analyzed. 

In the experimental evaluations presented in this work, we considered only the readout architecture with two fully connected layers, which corresponds to the case $q = 2$ as defined in the original MRGNN paper.  


For further theoretical details on other components of the original model that are not essential for understanding the next sections, refer to the original paper.

\subsection {\bf Fairing algorithm}
\begin{figure}
\centering
    \scalebox{0.8}{\begin{tikzpicture}
	\begin{pgfonlayer}{nodelayer}
		\node  (4) at (-3.5, -0.25) {};
		\node  (5) at (-3.5, 2.25) {};
		\node  (6) at (-2.5, 3) {};
		\node  (7) at (-2.5, 0.5) {};
		\node  (16) at (-3.5, -1) {\small $(I-\lambda K)$};
		\node  (18) at (-0.75, -1) {\small $(I-\mu K)$};
		\node  (19) at (-5, -1) {};
		\node  (20) at (-5, -1.5) {};
		\node  (21) at (1, -1.5) {};
		\node  (22) at (1, -1) {};
		\node  (23) at (-3.25, 3.5) {\small shrinking layer};
		\node  (24) at (-0.25, 3.5) {\small unshrinking layer};
		\node  (25) at (-2.25, -2) {\small N times};
		\node  (26) at (-6, 1) {$(X, A)$};
		\node  (37) at (0, 1) {$\cdots$};
		\node  (38) at (-6, 0.25) {\scriptsize input};
		\node  (47) at (3.5, 1) { $((I-\lambda K)(I -\mu K))^N X$};
		\node  (48) at (-2, 1) {$\cdot$};
		\node  (49) at (3.5, 0.25) {\scriptsize output};
		\node  (50) at (0.75, 1) {=};
		\node  (51) at (3.75, -1) {\small $-1 < \mu < 0 <\lambda < 1$};
		\node  (52) at (-4.5, 1) {$\longrightarrow$};
		\node  (53) at (-1.5, -0.25) {};
		\node  (54) at (-1.5, 2.25) {};
		\node  (55) at (-0.5, 3) {};
		\node  (56) at (-0.5, 0.5) {};
	\end{pgfonlayer}
	\begin{pgfonlayer}{edgelayer}
		\draw (5.center) to (6.center);
		\draw (6.center) to (7.center);
		\draw (7.center) to (4.center);
		\draw (5.center) to (4.center);
		\draw (19.center) to (20.center);
		\draw (20.center) to (21.center);
		\draw (21.center) to (22.center);
		\draw (54.center) to (55.center);
		\draw (55.center) to (56.center);
		\draw (56.center) to (53.center);
		\draw (54.center) to (53.center);
	\end{pgfonlayer}
\end{tikzpicture}}
    \caption{Scheme of the implementation of the Fairing algorithm introduced in \cite{taubin1995smoothing} with a GNN-like structure.}
    \label{fig:taubin}
\end{figure}
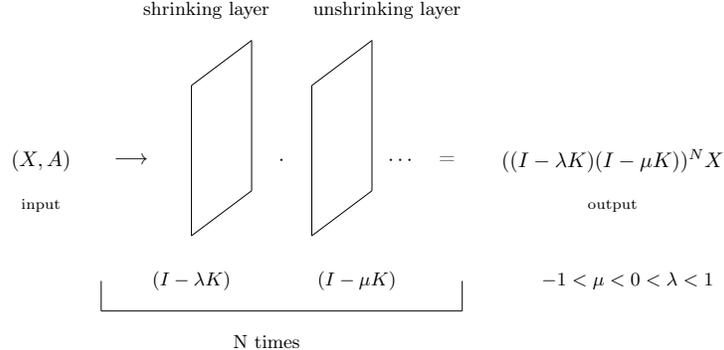

The Fairing algorithm, introduced in~\cite{taubin1995fair,taubin1995smoothing}, was proposed within the graph signal processing field. It aims to smooth graph signals without causing shrinkage and it is straightforward to see that it can be implemented in a GNN-like structure, as depicted in Fig.~\ref{fig:taubin}, where the symmetric operator $K$ defined in the paper acts as the signal propagator. The subsequent paragraphs will present an analysis of the algorithm using random walks to connect it with the reservoir computing framework discussed in~\cite{pasa2021multiresolution} and to give a spatial viewpoint on the spectral approaches.
As shown in Fig.~\ref{fig:taubin}, the core of the algorithm involves mathematically applying a degree-$2N$ polynomial of an operator $K$ to the initial embedding through iterative application of a degree-one polynomial of $K$. Parameters $\lambda$ and $\mu$ are tuned statically to ensure the transformation acts as a pass-band filter w.r.t. the frequencies of $K$, a symmetric operator with a spectrum bounded in $[0, 2]$. This is achieved by alternating a shrinking layer, obtained with the positive-valued parameter $\lambda$, with an unshrinking layer, obtained with the negative-valued parameter $\mu$. For further theoretical background refer to~\cite{taubin1995fair,taubin1995smoothing}. 
  
\section{Linking the Fairing algorithm with spectral GNN} 
In this section, we give our first contribution. We show how the Fairing algorithm can be linked with spectral GNN (MRGNN belongs to this class of architectures)
via the Dirichlet energy induced by the symmetric Laplacian matrix $\Delta_{\text{sym}}$,
defined as 
    \[
             \mathcal{E}^{\Delta_{\text{sym}}}(X) = \text{tr}(X^{\top}\Delta_{\text{sym}} X),
             \]
where  tr() is the trace operator. Consider now the gradient of $\mathcal{E}^{\Delta_{\text{sym}}}$, for simplicity in a graph scalar signal \mbox{$\vec{x} \in \mathbb{R}^n$:}
\[
\nabla_{\vec{x}}\mathcal{E}^{\Delta_{\text{sym}}}(\vec{x})= \nabla_{\vec{x}} \text{tr}\bigr(\vec{x}^{\top} \Delta_{\text{sym}} \vec{x}\bigl)=\nabla_{\vec{x}}\bigr(\vec{x}^{\top} \Delta_{\text{sym}} \vec{x}\bigl)
=\bigr(\Delta_{\text{sym}}^{\top} + \Delta_{\text{sym}}\bigl)\vec{x} = 2\Delta_{\text{sym}} \vec{x},
\]
i.e., $\mathcal{E}^{\Delta_{\text{sym}}}$ is a quadratic form induced by the symmetric Laplacian. In particular, it holds that its Hessian matrix is $2\Delta_{\text{sym}}$, that is positive semi-definite, so $\mathcal{E}^{\Delta_{\text{sym}}}$ is convex. This means that the following equality holds:
\[
\vec{x}' = (I+\eta\Delta_{\text{sym}})\vec{x}=\vec{x}+\frac{\eta}{2}\nabla_{\vec{x}}\mathcal{E}^{\Delta_{\text{sym}}}(\vec{x})
\]
i.e., a general Fairing layer can be interpreted as evolving graph signals along the gradient of $\mathcal{E}^{\Delta_{\text{sym}}}$, with direction determined by the sign of the hyperparameter $\eta$. This allows to link the Fairing algorithm with a spectral GNNs perspective: it is like to define a GNN that alternates gradient descent with gradient ascent using $\mathcal{E}^{\Delta_{\text{sym}}}$ as a loss function. The direction of the steps is defined by the sign of the coefficients, in analogy with the learning rate. If the steps in the negative direction of the gradient of  $\nabla\mathcal{E}^{\Delta_{\text{sym}}}$ are enough, then the composition will reduce $\mathcal{E}$.

\section{Tottering in MRGNN}

The main definition of the reservoir layer proposed in MRGNN~\cite{pasa2021multiresolution} is 
$ X' = \tanh(\Delta_{\text{sym}} X) $, where the activation function 
$ \tanh $ is introduced to mitigate the problem of \textit{tottering}. 
This phenomenon arises when random walks repeatedly traverse the same edge 
back and forth between two neighboring nodes, leading to redundant or 
uninformative paths.
The effect of the $\tanh $ function is to compress the signal values within 
the range $(-1, 1)$ through a smooth non-linearity. However, this operation 
uniformly shrinks all feature dimensions, without distinction. Moreover, 
since the reservoir layer is not subject to training, the inclusion of a 
non-linearity is not motivated by the usual aim of enlarging the hypothesis 
space toward non-linear functions.
To more precisely address redundancy while preserving the spectral nature of 
the original formulation, we propose an alternative method that allows for a 
more targeted influence on the tottering effect. A formal analysis of the 
redundancies induced by the $ \Delta_{\text{sym}} $ propagator will follow.
Consider the $k$-th power of the symmetric Laplacian:
\[
\Delta_{\text{sym}}^k=(I - D^{-1/2}A D^{-1/2})^k = (I - D^{-1/2}(AD^{-1})D^{1/2})^k 
\]
\[
=\sum_{t=0}^{k} \binom{k}{t} (-1)^t D^{-1/2}P^{\; t} D^{1/2} =   D^{-1/2} \biggl( \sum_{t=0}^{k} \binom{k}{t} (-1)^tP^{\; t} \biggr) D^{1/2}
\]
where $P=AD^{-1}$ is called \textit{transition matrix}:
\[
[P]_{ij} = \sum_{r } [A]_{ir}[D^{-1}]_{rj} =  \sum_{r} a_{ir}\frac{1}{d_{r}}\delta_{rj}=  \frac{a_{ij}}{d_{j}} = p(i|j) =
\begin{cases} 
\frac{1}{d_{j}} & \text{if } i \in \mathcal{N}_j, \\
0 & \text{otherwise}.
\end{cases}
\]
i.e., $[P]_{ij}$ is the probability of reaching node $i$ with a path of length $1$ starting from node $j$ when the probability to reach any first neighbour is uniform.\\
Let's see what happens for $t=2$:
\[
[P^{\; 2}]_{ij} = \sum_{r} [P]_{ir}[P]_{rj} =
\begin{cases} 
\sum_{r} p(i|r)p(r|j) \neq 0 & \text{if } \exists \; r \in \mathcal{N}_{i} \cap  \mathcal{N}_{j}, \\
0 & \text{otherwise}.
\end{cases}
\]
This means that the matrix elements of $P^{\;2}$ represents the probability of reaching node $i$ with a path of length $2$ starting from node $j$.\\ 
By repeated application, it's straightforward to see that in general $[P^{\;t+1}]_{ij}=\sum_{r} [P]_{ir}[P^{\;t}]_{rj}=\sum_{r} p(i|r)p_t(r|j)$ represents the probability to reach any node $r$ starting from node $j$ with a path of length $t$, multiplied by the probability to end in $i$ with one additional step.\\

The problem of tottering consists in the fact that random walks can travel edges going back and forth between two neighbors nodes indefinitely. For example, going from a node $j$ to node $i$ then come back to $j$ and finally going to node $i$, constitutes a path of length $3$ from $j$ to $i$.\\
Considering $[P^{\; 3}]_{ij}$, this specific cases contribute with the following terms:
\[
\sum_{r} \sum_l [P]_{ir}[P]_{rl}[P]_{lj} \delta_{il} \delta_{rj}=\sum_{r}\sum_{l} p(i|r)p(r|l)p(l|j)\delta_{il} \delta_{rj}= p(i|j)p(j|i)p(i|j) .
\]

Let's now analyze the binomial coefficients. One possible interpretation is to consider a random walk of length \textit{at most $k$} as a walk of length exactly $k$ if vacuous steps are allowed: at each step it can be decided to walk from a node to one of its first neighbors, or to stay still. \\
There are the following possibilities to reach $i$-th node with at most $k$ steps of which $t$ non-vacuous steps:
\begin{itemize}
\item $t=0 \rightarrow P^{\;0}=I$: there is just one way, i.e., to stand in $i$-th node;
\item $t=1 \rightarrow P$: to reach node $i$ when $j$-th node is a first neighbor, there are $k$ ways to chose when to do the single non-vacuous step;
\item $1<t\le k \rightarrow P^{\; t}$: there exist $\binom{k}{t}$ possibilities to select the $t$ necessary steps to reach the $i$-th node, provided it is reachable, and in the remaining $(k-t)$ steps it will be sufficient to stay still.
\end{itemize}
The sign, i.e., the factor $(-1)^t$, can be seen as an analogue of a phase to the contributions of the walks to the final signal on the $i$-th node after $k$ iterations, since $\Delta_{\text{sym}}$ propagates differences, then the contributions coming from first neighbours have opposite sign. And propagating \textit{differences of differences} preserves the sign, and therefore signals that reach $i$-th node with even non vacuous steps give a positive contribution, while for odd number of steps the phase is opposite, giving a negative contribution. Recall the expression:
\[
\Delta_{\text{sym}}^k
=\sum_{t=0}^{k} \binom{k}{t} (-1)^t D^{-1/2}P^{\; t} D^{1/2} 
= D^{-1/2} \sum_{t=0}^{k} \binom{k}{t} (-1)^tP^{\; t} D^{1/2} 
\]
Consider the operator in the middle (without considering the symmetrization applied by $D^{\pm1/2}$): for $k$ iterations, if this operator would be applied to a signal $X^{(0)}$, the resulting signal component $[X^{(k)}]_{ij}$ would be 
\[
[X^{(k)}]_{ij}=[(I-P)^{k}X^{(0)}]_{ij} = \sum_{r}[(I-P)^{k}]_{ir}x^{(0)}_{rj}
\]
where
\[
[(I-P)^{k}]_{ir} = \delta_{ir} +  \sum_{t=1}^{k} \binom{k}{t} (-1)^t p_{t}(i|r)
\]
and then
\[
[X^{(k)}]_{ij}=x_{ij}^{(0)}+  
\]
\[+\sum_{r \in \mathcal{V}}\biggl(- \binom{k}{1} p(i|r)  + \binom{k}{2} p_2(i|r) + \dots +(-1)^k\binom{k}{k}p_k(i|r) \biggr)x_{r j}^{(0)} .
\]
This formally shows that the $j$-th component of the signal on the $i$-th node after $k$ applications of $\Delta_{\text{sym}}$ results in a weighted average of the $j$-th component of the signal of the $k$-hop neighbours, where the weight assigned to the contribution from the $r$-th node depends on the total number of \textit{signed} random walks of length at most $k$ that exist between $r$-th node and $i$-th node. \\
Tottering occurs whenever a reflection between two nodes takes place. As a result, paths of length $t$ can be counted multiple times, depending on how many such back-and-forth steps (i.e., reflections) can be inserted. Since reflections preserve the parity of the path length, tottering amplifies the contributions of many paths. However, it does not affect paths that reach nodes whose minimum distance from the source is $k$ or $k-1$, as such paths cannot admit reflections without exceeding the length constraint.\\
The earlier expression shows that the signal from the $r$-th node is influenced by all possible random walks. Redundancies arise first from tottering walks, represented mathematically by powers of the $P$ matrix, and second from counting each walk with a multiplicity reflecting vacuous steps that leave the path unchanged, given by the $I$ matrix.
Suppose now to tune the weights with which this two redundant contributions are considered. One way to do that is to introduce a weighting parameter $\alpha$ to balance them, obtaining the following variant of the layer definition:
\[
X^{(k+1)}
=D^{-1/2}\big(\alpha I +(1-\alpha)P\big)^kD^{1/2}X^{(0)}
\]
the term $\big(\alpha I +(1-\alpha)P\big)^k$ consists in the following transformation for $x^{(t)}_{ij}$
\[
[X^{(t)}]_{ij}=\alpha^k x_{ij}^{(0)}+  
\]
\[
+\sum_{r \in \mathcal{V}}\biggl( \binom{k}{1} \alpha^{k-1}(1-\alpha) p(i|r)  + \binom{k}{2} \alpha^{k-2}(1-\alpha)^{2} p_2(i|r) +\cdots +
\]
\[
+\binom{k}{t} \alpha^{k-t}(1-\alpha)^{t}\alpha p_t(i|r) + \dots + \binom{k}{k}(1-\alpha)^k p_k(i|r) \biggr)x_{r j}^{(0)} .
\]
Rearranging the expression of the variant of the layer just analyzed:
\[
L = \alpha I +(1-\alpha)D^{-1/2}PD^{1/2}=I -(1-\alpha)\Delta_{\text{sym}}
\]
that for $0 < \alpha < 1$ coincides with the expression of an iteration of a Fairing layer as in \cite{taubin1995smoothing} where $\lambda = 1-\alpha$, i.e., a shrinking step. Therefore, a way to interpret the shrinking step from the point of view of random walks, is to read $\lambda$ as a tunable parameter that balances the redundancies, and since in this case $1-\alpha >0$, it does not account for the parity of walks.
The case of negative coefficients (that is, the unshrinking step) in the range $-1 < \mu < 0 < \lambda < 1$ can be expressed by considering $1< \alpha <2$ and recognizing $1-\alpha = \mu$, that instead accounts for the walks' parity.\\
Hence, tuning the Fairing parameters is equivalent to controlling the influence of random walks in an interpretable manner, in contrast to the effect of the $\tanh$ function, which instead shrinks all entries indiscriminately.
Based on this conclusions, it may be worthwhile to evaluate the impact of the Fairing algorithm on the reservoir component of MRGNN.

\section{Reservoir based on the Fairing Algorithm} 

In spatial terms, the Fairing algorithm allows for controlling the contribution of different random walks. Suppose we perform a static tuning of the parameter $\alpha$, meaning that its value is chosen analytically by the user (i.e., not learned empirically) to achieve a desired behavior. The effect of such tuning becomes immediately clear.
If we set $\alpha \sim \varepsilon$, with $0 <\varepsilon \ll 1 $, we suppress the contributions of random walks that contain any vacuous steps. In practice, this dampens all walks whose effective length is shorter than the maximum walk length $k$, leaving only the paths of length exactly $k$ to dominate the dynamics. However, among these, tottering walks have an high contribution, leading to the removal of one form of redundancy while leaving the other unchanged.
Conversely, if we choose $\alpha \sim 1-\varepsilon$, the contribution of all random walks is practically eliminated, and the propagator collapses to the identity function. This results in a model with limited expressivity.
A more balanced setting, such as $\alpha \sim 0.5$, mitigates both types of redundancy more equally. This observation may offer an explanation for why values of $\alpha$ in this range are also those empirically recommended in \cite{taubin1995smoothing} both for the shrinking and the unshrinking steps, as yielding fair smoothing without shrinkage.\\
The introduction of the Fairing algorithm is particularly suited for the basic definition of MRGNN, since $\Delta_{\text{sym}}$ has a limited spectrum $\sigma_{\Delta_{\text{sym}}} \subset [0, 2]$ exactly like the $K$ operator presented in \cite{taubin1995smoothing}. This suggests to exploit values of the constants $\lambda$, $\mu$ recommended in the paper to obtain an effect of smoothing without shrinkage.
The reservoir is now defined alternating a shrinking layer $X'= (I-\lambda{\Delta_{\text{sym}}})X $ with an unshrinking layer $X'= (I-\mu{\Delta_{\text{sym}}})X $, obtaining after $k$ iterations the following embedding matrix for a graph:
\[
X^{(k)}= (I-\lambda{\Delta_{\text{sym}}})^{n}(I-\mu{\Delta_{\text{sym}}})^{m}X^{(0)} 
\]
where $n+m = k$, since there are several combinations possible that allow to filter the signal in a different way.

\begin{figure}
    \captionsetup{justification=centering}
    \centering
    \includegraphics[width=0.7\textwidth]{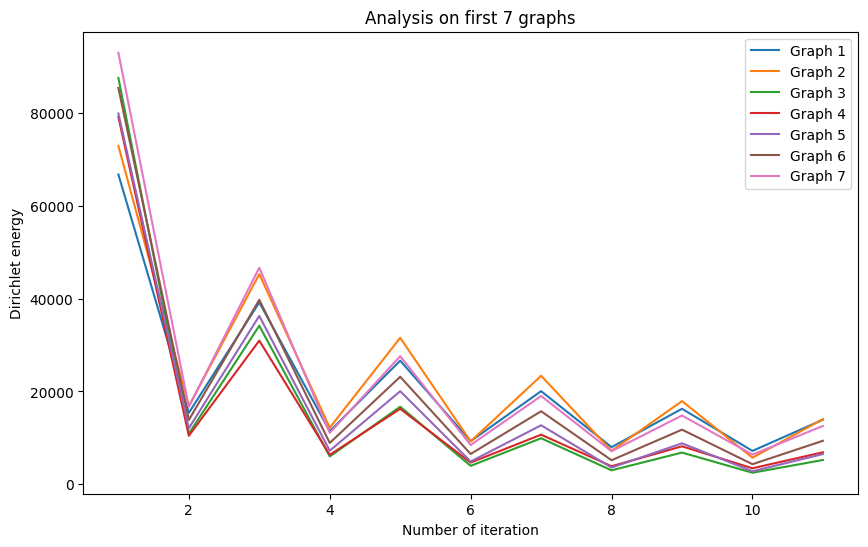}
    \caption{Plot of $\mathcal{E}^{\Delta_{\text{sym}}}$ for a Fairing reservoir $\lambda = 0.5$, $\mu=-0.66667$ of seven graphs of the dataset Enzymes after each iteration of the layer.}
    \label{L_symMULAM}
\end{figure}

\begin{figure}
    \captionsetup{justification=centering}
    \centering
    \includegraphics[width=0.7\textwidth]{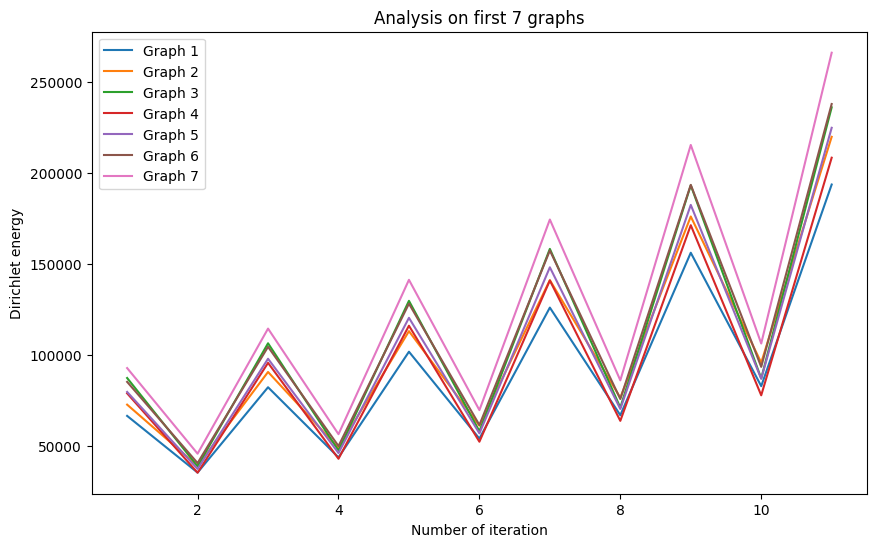}
    \caption{Plot of $\mathcal{E}^{\Delta_{\text{sym}}}$ for a Fairing reservoir $\lambda = 0.25$, $\mu=-0.5$ of seven graphs of the dataset Enzymes after each iteration of the layer.}
    \label{L_symMULAM2}
\end{figure}

The corresponding Dirichlet energy induced by $\Delta_{\text{sym}}$  can be written as  
\[
\mathcal{E}^{\Delta_{\text{sym}}}(X^{(k)})= \sum_{\omega \in \sigma_{\Delta_{\text{sym}}}} (1-\lambda\omega)^{2n}(1-\mu\omega)^{2m} \mathcal{E}^{\Delta^{\text{sym}}}_{\omega}(X^{(0)})
\]
that constitutes the filter reported in \cite{taubin1995fair}.
Depending on the values of the hyperparameters $\lambda$ and $\mu$, the filter selects a pass-band. \\
In figures Fig.~\ref{L_symMULAM}-\ref{L_symMULAM2} we report examples of plots of the Dirichlet energy 
for different values of $\lambda$ and $\mu$ for some of the graphs in the Enzymes dataset. 

In Fig.~\ref{L_symMULAM} it's clear the effect of the shrinking step, when the energy decreases, and the unshrinking step, when it increases. This perfectly represents the mathematical expectations, indeed, indicating a general step with $(I-\alpha \Delta_{\text{sym}})$ :
\[
\mathcal{E}^{\Delta_{\text{sym}}}(X^{(t+1)})= \sum_{\omega \in \sigma_{\Delta_{\text{sym}}}} (1-\alpha\omega)^2 \mathcal{E}^{\Delta^{\text{sym}}}_{\omega}(X^{(t)})
\]
\[
=\mathcal{E}^{\Delta_{\text{sym}}}(X^{(t)})+\sum_{\omega \in \sigma_{\Delta_{\text{sym}}}} (\alpha\omega-2)\alpha\omega\mathcal{E}^{\Delta^{\text{sym}}}_{\omega}(X^{(t)}) 
\]
that for a shrinking step, i.e. $0< \alpha= \lambda < 1$ results in $(\alpha\omega-2)\alpha\omega < 0$, i.e., a decrease of the energy, while for $\alpha= \mu <0$ results in $(\alpha\omega-2)\alpha\omega > 0$, that is, an increase of the energy, reflecting the effect of unshrinking.

Acting on the hyperparameters $\lambda$ and $\mu$ allows to tune the frequencies that are filtered: decreasing $\lambda$ to $0.25$ and $\mu$ to $-0.5$ causes a shift of the pass-band treshold $k_{PB}= \frac{1}{\lambda}+\frac{1}{\mu}$ to $2$, making the resulting filter passing the complete band, with the resulting effect of amplification after many iterations (see Fig.~\ref{L_symMULAM2}). 

This methodology allows to control the signal’s energy by adjusting the hyperparameters of the filter. It is particularly interesting because it is based on the same mathematical principles as GCNs, even though it was published in 1995, slightly before the widespread research on GNNs. This highlights that the side effects of smoothing were already understood and managed before the rise of GNNs. 

\section{Exploratory experiments}

A limited experimental campaign was conducted 
to preliminarily assess the potential of the proposed approach, with no aim to perform a fair comparison versus the MRGNN model. Specifically,
we reused the hyperparameters already optimized on the MRGNN model, observing encouraging results.
This suggests that the modified reservoir architecture is promising, and that a more thorough hyperparameter tuning could further enhance its competitiveness.\\

\subsection {Datasets} 
Experiments on the proposed reservoir were conducted on four molecular graph classification benchmarks: PTC \cite{helma2001}, NCI1 \cite{wale2008}, PROTEINS \cite{borgwardt2005}, and ENZYMES \cite{borgwardt2005}. These datasets were selected to ensure comparability with the original MRGNN model, which was tested on the same benchmarks.\\
As described in \cite{pasa2021multiresolution}, PTC and NCI1 consist of molecular graphs representing chemical compounds, where nodes correspond to atoms and edges to chemical bonds. The classification task in PTC aims to determine the carcinogenicity of compounds in male rats, while in NCI1 the goal is to classify compounds based on their activity in anticancer screens. PROTEINS and ENZYMES are datasets in which graphs represent proteins, with nodes denoting amino acids and edges connecting amino acids that lie within a 6Å distance in the 3D structure. ENZYMES involves a six-class classification task, whereas the others are binary.
\begin{table}[t]
\centering
\small
\caption{Molecular graph datasets statistics.}
\label{tab:molecular_datasets}
\begin{center}
\begin{tabular}{|l|c|c|c|c|c|}
\hline
\textbf{Dataset} & \textbf{\#Graphs} & \textbf{\#Nodes} & \textbf{\#Edges} & \textbf{Avg \#Nodes} & \textbf{Avg \#Edges} \\
                 &                   &                  &                  & \textbf{per Graph}    & \textbf{per Graph}   \\
\hline
\textbf{PTC}     & 344               & 4915             & 10108            & 14.29                 & 14.69                \\
\textbf{NCI1}    & 4110              & 122747           & 265506           & 29.87                 & 32.30                \\        
\textbf{PROTEINS}& 1113              & 43471            & 162088           & 39.06                 & 72.82                \\
\textbf{ENZYMES} & 600               & 19580            & 74564            & 32.63                 & 124.27               \\
\hline
\end{tabular}
\end{center}
\end{table}

\subsection {Implementation details}
The implementation of the Fairing variant of the reservoir was implemented modifying the original code of MRGNN publicly available on GitHub\footnote{https://github.com/lpasa/MRGNN}.
In particular, a variant of the reservoir was defined with a new reservoir version that, depending on the parity of the reservoir layer iteration, applies either a shrinking step ($\lambda = 0.5$) or an unshrinking step ($\mu = -0.66667$).

\subsection{Results}
In order to test the impact of the Fairing algorithm on the reservoir definition, experiments were conducted for each dataset using the ``optimal'' values of the hyperparameters documented in the MRGNN original paper (reported in Table~\ref{tab:hyperparams}). Then, a 10-fold cross-validation was done for 5 runs. For each split of the 10-fold cross-validation, the highest value of the validation accuracy was selected, and it was selected the test accuracy of the epoch nearest to the last one to which the highest value of the validation accuracy was associated. In this way, for each run expectation values of the test accuracies were calculated, and finally, the expectation of this expectations was calculated with respect to the five runs. The obtained results are reported in Table~\ref{tab:results}, jointly with the ones obtained by MRGNN so to provide an  idea of the performance  of the original architecture. These very preliminary Fairing-based reservoir results exhibit values that are comparable to those obtained by MRGNN after a complete model selection,  suggesting that a comparable model selection procedure for the proposed approach could lead to SOTA results.


\begin{table}[t]
\centering
\scriptsize
\caption{Hyperparameters used in the experimental setup for each dataset.}
\label{tab:hyperparams}
\begin{center}
\begin{tabular}{|l|c|c|c|c|c|c|c|}
\hline
\textbf{Dataset} & \textbf{epochs} & \textbf{LR} & \textbf{Dropout} & \textbf{Weight Decay} & \textbf{Batch Size} & \textbf{\# Hidden}  & \textbf{k} \\
\hline
PTC      & 400 & 0.0005 & 0.6  & 0.005   & 32 & 15  & 4 \\
NCI1     & 500  & 0.001 & 0.5  & 0.0005   & 32 & 100  & 4 \\
PROTEINS & 200  & 0.001  & 0.5  & 0.0005  & 32 & 50  & 5 \\
ENZYMES  & 200  & 0.001  & 0.5  & 0.0005   & 32 & 100  & 6 \\
\hline
\end{tabular}
\end{center}
\end{table}

\begin{table}[t]
\caption{Mean test accuracies for MRGNN (Table III-\cite{pasa2021multiresolution}) and the Fairing approach. Results for the Fairing approach are: {\it i)} obtained by using the optimal hyper-parameters's values of MRGNN ({\bf no model selection performed}); {\it ii)} computed over five runs of 10-folds cross-validation, with sample standard deviation defined as $s = \sqrt{s^2}$, where
$s^2 = \frac{1}{n - 1} \sum_{i=1}^{n} (x_i - \bar{x})^2$, $\bar{x}$ is the sample mean, and $n$ denotes the number of runs. 
\label{tab:results}}
\centering
\small
\begin{center}
\begin{tabular}{|l|c|c|}
\hline
\textbf{Dataset} &  \textbf{MRGNN Test Accuracy (\%)} & 
\textbf{Fairing Test Accuracy (\%)} \\
\hline
PTC    & 57.60 $\pm$ 10.01  &  59.58 $\pm$ 1.73 \\
NCI1 & 80.58 $\pm$ 1.88 & 79.02 $\pm$ 0.27 \\ 
PROTEINS  & 75.84 $\pm$ 3.51 & 74.39 $\pm$ 0.83 \\
ENZYMES   & 68.20 $\pm$ 6.86 & 68.37 $\pm$ 1.89 \\
\hline
\end{tabular}
\end{center}
\begin{flushleft}
\end{flushleft}
\end{table}

\section{Conclusions}
This work introduced a theoretical analysis of the impact of Fairing algorithms when applied to reservoir computing architectures. In particular, it derived and demonstrated the connection between the smoothing behavior induced by Taubin Fairing and the dynamics of random walks on graphs, showing how such Fairing acts on the structural and spectral properties of the underlying reservoir. This theoretical contribution offers a perspective on the role of graph-based smoothing in reservoir dynamics and lays the groundwork for further extensions.\\
For what concerns the exploratory experiments proposed, while no formal comparison was performed, the observed similarity in performance suggests that a more rigorous evaluation between the original and Fairing-based models could be an interesting direction for future research.

\end{document}